# Peeking Inside the Schufa Blackbox: Explaining the German Housing Scoring System

**Dean-Robin Kern**
University of Siegen
Germany
dean-robin.kern@uni-siegen.de

**Gunnar Stevens**
University of Siegen
Germany
gunnar.stevens@uni-siegen.de

**Erik Dethier**
Hochschule Bonn-Rhein-Sieg
Germany
erik.dethier@h-brs.de

**Sidra Naveed**
University of Siegen
Germany
sidra.naveed@uni-siegen.de

**Fatemeh Alizadeh**
University of Siegen
Germany
fatemeh.alizadeh@uni-siegen.de

**Delong Du**
University of Siegen
Germany
delong.du@student.uni-siegen.de

**Md Shajalal**
University of Siegen
Fraunhofer FIT, Germany
md.shajalal@fit.fraunhofer.de

## Abstract
Explainable Artificial Intelligence (XAI) is a concept aimed at making complex algorithms transparent to users through a uniform solution. Researchers have highlighted the importance of integrating domain-specific contexts to develop explanations tailored to end-users. In this study, we focus on the Schufa housing scoring system in Germany and investigate how users' information needs and expectations for explanations vary based on their roles. Using the speculative design approach, we asked business information students to imagine user interfaces that provide housing credit score explanations from the perspectives of both tenants and landlords. Our preliminary findings suggest that although there are general needs that apply to all users, there are also conflicting needs that depend on the practical realities of their roles and how credit scores affect them. We contribute to Human-centered XAI (HCXAI) research by proposing future research directions that examine users' explanatory needs considering their roles and agencies.

intelligence;

## Introduction
Credit and consumer scoring systems play a crucial role in determining individuals' eligibility for credit, services, and rentals, significantly affecting their lives [28, 25, 1]. The integration of complex machine learning algorithms into these systems has led to them being regarded as "black boxes" with unclear computational processes [17, 8, 13, 24, 27, 7, 15, 18, 22, 23, 14]. This lack of transparency, coupled with potential biases and the right to explanations, has led to calls for increased transparency and explainability in these high-risk systems [13, 16, 23, 5, 12].

A notable instance of such an opaque scoring system in Germany is the Schufa scoring system. As a credit reporting agency, Schufa monitors all bills and fines over time to establish and maintain credit scores for German residents. Residents can access their Schufa records, which include a score ranging from 0 to 100, indicating their creditworthiness, for a small fee (See Figure 1) [1].

**Figure 1:** The risk associated with the Schufa score [6]

The Schufa record impacts a wide range of aspects in the lives of German residents, with approximately 10,000 partner companies from various industries, such as finance, insurance, telecommunications, and retail [1]. In the housing context, landlords use the Schufa score to evaluate the creditworthiness of potential tenants and choose the most suitable candidate from a large pool of applicants. Despite its extensive use, the agency's scoring algorithms remain opaque, and the calculation of scores is undisclosed [25]. An investigation by AlgorithmWatch and Spiegel Online/Bayerische Rundfunk has revealed possible biases in the Schufa system [3], emphasizing the need for a more transparent and explainable scoring system.

In response to the issues surrounding Schufa's opacity, the OpenSCHUFA [2] initiative has emerged as a prominent example of efforts to improve transparency and fairness in the credit scoring system. OpenSCHUFA aims to foster public scrutiny by collecting and analyzing anonymized Schufa data voluntarily provided by individuals. By crowdsourcing data and conducting independent analysis, the initiative seeks to uncover potential biases, improve public understanding of the scoring system, and ultimately contribute to the development of a more transparent and equitable credit scoring process.

Research indicates that different user groups have varying explanatory needs [26]. Identifying these diverse needs across user groups involved in scoring systems is a significant challenge, given the numerous stakeholders. This study expands on previous research by examining how users' explanatory needs differ based on their relationship with the Schufa housing scoring system in Germany. Our research questions are: (a) what explanation needs do users have based on their interactions with the system, and (b) how do these needs differ across user groups? We employed a speculative design approach, tasking students with identifying potential explanatory needs for housing credit scores from the perspectives of tenants and landlords. The preliminary findings of this study could contribute to the development.

## Explaining credit scoring systems
Designing clear and understandable explanations for credit scoring systems is essential due to their increasing use in financial decision-making. Credit scores play a critical role in determining access to financial services and opportunities. Ensuring that decisions based on these scores are fair and transparent is crucial. Transparent explanations for credit scoring systems can promote accountability and trust in the financial system while mitigating potential biases and discrimination against certain individuals or groups. Providing explanations tailored to users' needs ensures fairness, transparency, and accountability in financial decision-making.

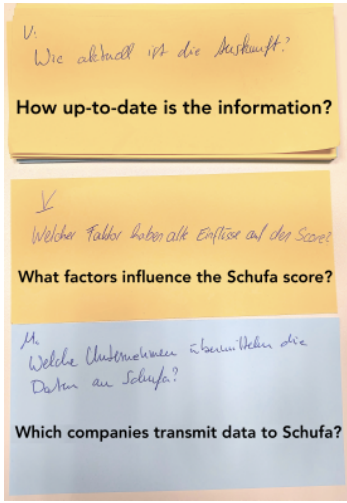

**Figure 2:** The picture shows three German cards with handwritten text, featuring questions from the perspectives of tenants and landlords, along with their corresponding English translations.

To meet the need for greater transparency in complex machine learning algorithms, several studies (e.g., [5, 13, 24, 20, 21, 19]) have focused on designing explanations for credit scoring systems. Demajo et al., for instance, developed an accurate and interpretable credit scoring model using the XGBoost algorithm, which they augmented with a 360-degree explanation framework. This framework provides different types of explanations tailored to various needs. The authors found it to be simple, consistent, and reliable, highlighting the importance of interpretability in credit scoring [13].

Various eXplainable AI (XAI) methods, such as SHAP+GIRP, Anchors, ProtoDash, LIME, DeepLIFT, and SHAP, have been explored as potential solutions in previous works [13, 24]. In addition, researchers have investigated the optimization of economic target functions and the use of counterfactual explanations, considering human factors [9, 21, 19, 10, 18]. For example, Chromik conducted a study on the accessibility of XAI for non-proficient users, analyzing human factors and cognitive biases that influence users' comprehension of XAI explanations and presenting case studies to support his findings [10].

Although the recognition of Human-Centric eXplainable AI (HCXAI) has been emerging, such as cognitive biases and user preferences for explanations [26], the agencies of users' roles and pragmatics utilizing explanations are often neglected in research [10, 15, 22]. A particular research gap lies in the insufficient investigation of context-specific explanation requirements for diverse user parties, such as landlords or tenants. The influence of users' roles on specific explanatory needs and its consequent impact on design remains indeterminate.

## Methodology

In our preliminary investigation, we conducted a speculative design workshop [4] that employed a forward-looking approach to imagine potential future scenarios, challenge preconceptions, and raise questions regarding the Schufa score and XAI systems. Our speculative workshop aimed to provoke critical thinking and stimulate a dialogue about the complex implications of explainable AI in credit scoring systems.

We invited a group of eight business information systems and design students, without any specific recruitment criteria, to participate in this workshop. Before the workshop, the participants were provided with an overview of the Schufa score simulator, the Schufa principle, and XAI concepts to establish a common understanding of the context. We then divided the participants into two groups: "The Landlord" and "The Tenant". Each group was asked to generate queries that an imaginary Schufa score XAI system should answer, with the aim of identifying the information and explanation needs of participants based on their assigned roles.

The workshop lasted two lecture lengths (approximately 180 minutes), and consent was obtained for the scientific processing of the workshop materials. 49 questions were formulated, with 18 related to the tenant's perspective and 31 to the landlord's perspective (see Figure 2). We used thematic clustering proposed by Brown and Clarke [11] to analyze the data and identify significant themes, patterns, and insights regarding explanatory needs. By adopting the speculative design approach, the workshop encouraged participants to explore and reflect on broader issues, chal-lenges, and possibilities, paving the way for more informed and innovative solutions in HCXAI research.

## Preliminary Findings and Analysis

The preliminary findings indicate three categories of HCXAI explanatory needs, namely (a) consistent needs that are tenant and landlord's common interests, (b) distinct needs that are unique to each role, and (c) conflicting needs between two roles.

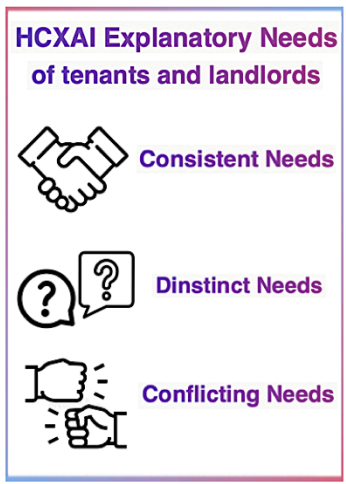

**Figure 3:** Three categories of HCXAI explanatory needs of tenants and landlords

**Consistent Needs**: Our preliminary findings revealed three consistent explanatory needs that were shared by tenants and landlords: 1) Timeliness, 2) Error identification, and 3) Transparency. Participants' queries related to the Timeliness focused on the frequency of updating and recalculation of the Schufa score. Examples inquiries from both parties include: *Is the Schufa Score updated monthly? How often is the Schufa Score recalculated and how up-to-date is it?* Both parties posed questions regarding errors in the Schufa Score. Both parties were interested in how to identify potential errors, such as: *Do companies transmit my data to Schufa without errors? How can I correct incorrect data? How do I recognize errors? What happens if the algorithm makes a mistake? What about incorrect values?* In addition, both parties also inquired about the transparency of the process, specifically regarding i) the data source (*Where does Schufa obtain its data from?*), ii) the algorithm (*On what formula is the Schufa score based?*), iii) the input factors (*What factors impact the Schufa score?*), and iv) the Schufa Score services (*How current is the information? What if two people want to live in a household? Schufa score for both?*).

**Distinct Needs**: The distinct needs of both parties in interpreting the Schufa score became evident, with landlords seeking more information on the factors that affect the score and its accuracy (*What is the accuracy of the scores?*), while tenants were more focused on understanding how to maintain a good score (*How do I achieve a good Schufa score?*). Furthermore, landlords were increasingly interested in the use of the Schufa score as a tool to make informed decisions *(Landlord satisfaction? How do tenant problems affect the Schufa score?)* as well as the legal aspects (*Can Schufa be sued for incorrect information?*). While tenants were more interested in modifications (*Can I delete entries?*). This indicates that the information that users seek varies depending on their perspective and role.

**Conflicting Needs**: Ethical concerns regarding privacy protection and bias in the Schufa house score system have led to conflicting needs for tenants and landlords. Tenants are interested in obtaining a favorable Schufa score without having to provide detailed personal information, while landlords prioritize accuracy in the Schufa score. This conflict may also be linked to data privacy concerns. To elaborate, tenants may be concerned about the amount and type of personal data used to calculate their Schufa score. This concern can be addressed with questions such as: (Can I delete my Schufa entries? Who has access to my Schufa score and data?). On the other hand, landlords may be primarily concerned with how the system can be utilized and its reliability (Financial obligations? Family situation? Previous tenancy? Information about tenant's employment. What is the monthly income? Current expenses of the tenant? Asset situation? Does the tenant have any legal problems?).

The relevance of the identified different needs - consistent, distinct, and conflicting - applies to the HCXAI domain beyond just the housing market. Addressing these needs in HCXAI systems promotes better understanding and trust in AI-driven decision-making across various sectors.

Consistent needs, such as timeliness, error identification, and transparency, are fundamental requirements for any AI system to be effective and trustworthy across different industries.

Distinct needs emphasize the importance of providing role-specific information in HCXAI systems, ensuring that they cater to the unique requirements of different user agencies in various applications.

Conflicting needs highlight the need for ethical considerations and privacy protection in HCXAI systems, ensuring that the rights and interests of all parties are respected in different contexts.

Overall, addressing these needs is highly relevant to the HCXAI domain, fostering more equitable, ethical, and reliable outcomes across various sectors and applications.

## Conclusion

The development and deployment of Human-Centered Explainable Artificial Intelligence (HCXAI) systems are crucial in addressing the diverse and often conflicting needs of different user agencies. In the context of the Schufa scoring system, both landlords and tenants demand transparency and interpretability, yet their distinct utilization requirements can lead to conflicts. It is imperative for credit and consumer scoring systems to provide explanations that adhere to data protection regulations, ensuring that no party is unfairly disadvantaged.

HCXAI systems play a vital role in offering pertinent and actionable explanations to each user agency while safeguarding the rights and interests of all involved parties. The potential value conflict arising from the demand for algorithmic transparency by tenants and landlords against the interests of the Schufa organization, who may be hesitant to disclose complete insight into their systems, underscores the complexity of balancing competing stakeholder interests. This conflict could stem from the organization's concerns that full transparency might allow for the manipulation of the system by tenants or landlords. Addressing such conflicts is crucial and should be a focal point in future research.

The necessity of considering the diverse needs of user agencies in HCXAI is paramount, as neglecting to do so could result in substantial challenges and hinder the effective and ethical deployment of these systems. Future work should prioritize conducting workshops with actual stakeholders, rather than solely relying on students engaging in role-playing exercises. Engaging with genuine stakeholders will enable researchers to delve deeper into the specific needs, concerns, and expectations of different user agencies, ultimately leading to the development of more robust, effective, and ethically sound HCXAI systems.